\documentclass[10pt,a4paper]{article}
\usepackage[latin1]{inputenc}
\usepackage{amsmath}
\usepackage{amsfonts}
\usepackage{amssymb}
\usepackage{graphicx}

\usepackage[latin1]{inputenc}
\usepackage{amsmath}
\usepackage{amsfonts}
\usepackage{amssymb}
\usepackage{graphicx}
\usepackage{marvosym}

\usepackage[table]{xcolor}
\usepackage{tabularx}

\usepackage[disable]{todonotes} 

\usepackage{amsmath}

\newcommand{\dif}{\textrm{d}}

\usepackage{subfigure}

\usepackage[export]{adjustbox}
\usepackage{graphicx}

\usepackage{multirow}

\begin{document}

\title{Gaussian Processes for Music Audio Modelling and Content Analysis}

\author{ Pablo A. Alvarado, Dan Stowell \\ Queen Mary University of London}

\maketitle

\begin{abstract}
\noindent
Real music signals are highly variable, yet they have strong statistical
structure. Prior information about the underlying physical mechanisms by which sounds are generated and rules by which complex
sound structure is constructed (notes, chords, a complete musical score), can
be naturally unified using Bayesian modelling techniques. Typically algorithms
for Automatic Music Transcription independently carry out individual tasks such
as multiple-F0 detection and beat tracking. The challenge remains to perform
joint estimation of all parameters. We present a Bayesian approach for modelling music audio, and content analysis.
The proposed methodology based on Gaussian processes seeks joint estimation of
multiple music concepts by incorporating into the kernel prior information
about non-stationary behaviour, dynamics, and rich spectral content present in the modelled music signal.
We illustrate the benefits of this approach via two tasks: pitch estimation, and  inferring missing segments in a polyphonic audio recording.
\end{abstract}

\section{Introduction}
%
\noindent
In music information research, the aim of audio content analysis is to estimate musical concepts which are present but hidden in the audio data \cite{serra13}. 
With this purpose, different signal processing techniques are applied to music signals for extracting useful information and descriptors related to the musical concepts. 
Here, musical concepts refers to parameters related to written music, such as pitch, melody, chords, onset, beat, tempo and rhythm.  
Then, perhaps the most general application is one which  involves the prediction of several musical dimensions, that of recovering the score of a music track given only the audio signal \cite{muller11}. This is known as automatic music transcription (AMT) \cite{benetos13}.

%
%
AMT refers to extraction of a human readable and interpretable description from a recording of a music performance. We refer to polyphonic AMT in cases where more than a single musical pitch plays at a given time instant. The general task of interest is to infer automatically a musical notation, such as the traditional western music notation, listing the pitch values of notes, corresponding timestamps and other expressive information in a given audio signal of a performance \cite{cemgil10}.
Transcribing polyphonic music is a nontrivial task, especially in its more unconstrained form when the task is performed on an arbitrary acoustical input, and music transcription remains a very challenging problem \cite{benetos12}.
%
	 
Real music signals are highly variable, but nevertheless they have strong
statistical structure. Prior information about the underlying structures, such as knowledge of the physical mechanisms by which sounds are generated, and knowledge  about the rules by which complex sound structure is compiled (notes, chords, a complete musical score), can be naturally unified using Bayesian hierarchical modelling techniques. This allows the formulation of highly structured probabilistic models \cite{cemgil10}. 
On the other hand, typically, algorithms for AMT are developed independently to carry out individual tasks
such as multiple-F0 detection, beat tracking and instrument recognition. The challenge remains
to combine these algorithms, to
perform joint estimation of all parameters \cite{benetos13}.

We present the design, implementation, and results of experiments of an alternative Bayesian approach for audio content analysis on monophonic, and polyphonic music signals with the possibility of being used for AMT. 
We use Gaussian process (GP) models for jointly uncovering music concepts from audio, by introducing a direct connection between the music concepts and the model hyper-parameters.
The proposed methodology allows to incorporate in the model prior information about physical or mechanistic behaviour,  nonstationarity, time dynamics (local periodicity, and non constant amplitude envelope), spectral harmonic content, and musical structure, latent in the modelled music signal. 
Specifically in the context of music informatics, we present kernels that embody a probabilistic model of music notes as time-limited harmonic signals with onsets and offsets. 
A comparison with related work is provided in section \ref{related_work}.
We illustrate the benefits of this approach via two tasks: pitch estimation, and  inferring missing segments in a polyphonic audio recording.
As part of working towards a high-resolution AMT system, out method correctly estimates polyphonic pitch while performing these sample-level tasks.

\section{Gaussian process regression for music signals}
\noindent
%
Gaussian process-based machine learning is a powerful Bayesian paradigm for nonparametric nonlinear regression and classification \cite{Sarkka2013}. 
Gaussian Processes (GPs) can be defined as distributions over functions such that any finite number of function evaluations $\textbf{f}=[f(t_1),\cdots, f(t_N)]$, have a jointly normal distribution \cite{rasmussen05}. A GP is completely specified by its mean function $\mu(t) = \mathbb{E}[f(t)]$ (in this work it is assumed to be $\mu(t)=0$),
%
%
and its kernel or covariance function 
\begin{align}\label{e.covgen}
k(t,t') = 
\mathbb{E}\left[ (f(t) - \mu(t)) ( f(t') - \mu(t'))\right] ,
\end{align} 
where $k(t,t')$ has hyper-parameters $\boldsymbol{\theta}$. 
We write the Gaussian process as
\begin{align}
f(t) \sim \mathcal{GP}(\mu(t),k(t,t')).
\end{align}
%
%
%
The regression problem concerns the prediction of a continuous quantity \cite{rasmussen05}, here a function $f(t)$, 
given a data set $\mathcal{D} = \left\lbrace (t_i,y_i) \right\rbrace_{i=1}^{N} $, where $y_i$ are assumed as noisy measurements of  $f(t)$ at typically regularly-spaced time instants $t_i$ (though GP regression framework allows for irregular sampling or missing data), i.e. 
$y_i = f(t_i) + \epsilon_i$,
%
%
%
where $\epsilon_i \sim \mathcal{N}(0,\sigma_{\text{noise}}^2)$.
In GP regression for mono channel audio signals, instead of estimating parameters $\boldsymbol{\eta}$ of fixed-form functions $f(t,\boldsymbol{\eta}): \mathbb{R} \mapsto \mathbb{R}$ where the time input variable $t \in \mathbb{R}$, we model the whole function $f(t)$ as a GP. That is, instead of putting a prior over the function parameters $\boldsymbol{\eta}$, we introduce a prior over the function $f(t)$ itself \cite{Sarkka2013b}. 
Learning in GP regression corresponds to computing the posterior distribution over the function $f(t)$ conditioned on the observed data $\textbf{y}=\left[y_1,\cdots,y_N \right]^{\top} $ \cite{Sarkka2011,Roberts12}.
The underlying idea in GP regression is that the correlation function introduces dependences between function $f(t)$ values  at different inputs. Thus, the function values at the
observed points give information also of the unobserved points \cite{Sarkka2013}.
The structure of the kernel \eqref{e.covgen} captures high-level properties of the unknown function $f(t)$, which in turn determines how the model generalizes or extrapolates to new test time instants \cite{lloyd2014}.
%
%
This is quite useful because we can introduce prior knowledge about what we believe the proprieties of music signals are, by choosing a proper kernel that reflects those characteristics. In section \ref{section:kernel_design} we study in more detail the design of kernels.

%
%
%


\subsection{Model definition}
%
%
%
Under a non-parametric Bayesian regression approach using Gaussian processes we are interested in calculating the posterior distribution over a stochastic function evaluated at test points $\textbf{t}_{*}$, that is, the joint distribution of the vector $\textbf{f}$ observed only via noisy measurements $\textbf{y}$, then
\begin{align}\label{e.model1}
p(\textbf{f}|\textbf{y}) = 
\frac{
p(\textbf{y}|\textbf{f})
\times
p(\textbf{f}|\boldsymbol{\theta})
}{
p(\textbf{y})
},
\end{align}
where $p(\textbf{y}|\textbf{f})$ corresponds to the likelihood, $p(\textbf{f}|\boldsymbol{\theta})$ to the prior, $\boldsymbol{\theta}$ are the model hyper-parameters (prior parameters), $p(\textbf{y})$ is the evidence or marginal-likelihood, and $p(\textbf{f}|\textbf{y})$ is the posterior or conditional predictive distribution. We describe each of these four expressions in the next sections.

\subsubsection{Likelihood}
%
%
%
%
Assuming that conditioned on the $f(t_i)$ the signal observations $y_i$ are i.i.d. (independent and identically distributed), then the joint probability distribution of all the observations $\textbf{y}$ follows a Gaussian distribution corresponding to
\begin{align}
p(\textbf{y}|\textbf{f},\sigma^{2}_{\text{noise}})
=  
\mathcal{N}(\textbf{y}|\textbf{f},\sigma^{2}_{\text{noise}}\textbf{I}_{N}),
\end{align}
where $f_i = f(t_i)$
%
and  $\textbf{I}_{N}$ is an identity matrix of size $N$.
%

%
\subsubsection{Prior}

Using the definition of GPs introduced at the beginning of this section, and
knowing that we have a finite set of corrupted observations $\textbf{y}$, then the finite set of GP function evaluation values $\mathbf{f}$ follows a normal marginal distribution $p(\textbf{f}|\boldsymbol{\theta})$ conditioned on the hyper-parameters $\boldsymbol{\theta}$, whose mean is zero and whose covariance is defined by a Gram matrix $\textbf{K}_f$, this is 
\begin{align}\label{e.prior_model1}
p(\textbf{f}|\boldsymbol{\theta}) =
\mathcal{N}(\textbf{f}|\boldsymbol{0}, \textbf{K}_f),
\end{align}
where the covariance matrix is calculated using \eqref{e.covgen}, i.e. $[\textbf{K}_f]_{i,j} = k(t_i,t_j)$ \cite{bishop06}.

\subsubsection{Marginal-Likelihood}
The \textit{marginal-likelihood} (or evidence) $p(\textbf{y})$ mentioned before in \eqref{e.model1} is  the integral of the likelihood times the prior \cite{rasmussen05}
\begin{align}\label{e.marlikelihood}
p(\textbf{y}) = \int p(\textbf{y}|\textbf{f}) p(\textbf{f}|\boldsymbol{\theta}) \dif \textbf{f}.
\end{align}
Since the likelihood $ p(\textbf{y}|\textbf{f})$ and the prior $p(\textbf{f}|\boldsymbol{\theta})$ are multivariate Gaussian distributions, we can calculate directly the integral in \eqref{e.marlikelihood}. Using the properties of the normal distribution \cite{bishop06} for marginal and conditional normal distributions, we obtain
%
\begin{align}\label{e.marlikelihood_model1}
%
p(\textbf{y}) =
\mathcal{N}(\textbf{y}|\boldsymbol{0},\textbf{K}_{y}),
\end{align}
%
where the values in the matrix $\textbf{K}_{y} =  \textbf{K}_{f} + \sigma^{2}_{\text{noise}}\textbf{I}$ depend on the hyper-parameters $\boldsymbol{\theta}$ (we have included $\sigma^{2}_{\text{noise}}$ in the hyper-parameters vector). The reason it is called the marginal likelihood, rather than just likelihood, is because we have marginalized out the latent Gaussian vector $\textbf{f}$ \cite{murphy12}.


\subsubsection{Posterior}
The computation of the
posterior distribution of the Gaussian process conditioned on
the set of measurements $\textbf{y}$ and estimation of the parameters $\boldsymbol{\theta}$ of the
covariance function of the process correspond to learning in
this non-parametric model \cite{Sarkka2013}.
Using the properties of Gaussian distribution \cite{bishop06,rasmussen05}, the posterior has the form
%
\begin{align}\label{e.posterior_model1}
%
p(\textbf{f}|\textbf{y}) =
\mathcal{N}(\textbf{y}|\boldsymbol{\mu}_{\text{pos}},\textbf{K}_{\text{pos}}),
\end{align}
%
where the posterior mean is $\boldsymbol{\mu}_{\text{pos}} =  \textbf{K}_{f}\textbf{K}_{y}^{-1}\textbf{f}$, and the posterior covariance matrix is $\textbf{K}_{\text{pos}} = \textbf{K}_{f} - \textbf{K}_{f}^{\top}\textbf{K}_{y}\textbf{K}_{f}$.

\subsection{Kernel design}\label{section:kernel_design}
The covariance function \eqref{e.covgen} used for computing the prior distribution \eqref{e.prior_model1} allows us to introduce in the model all the knowledge and beliefs we have about the properties of the data. We are trying to model music signals, and some of the broad properties of audio signals are non-stationarity, rich spectral content, dynamics (locally periodic, non constant amplitude envelope), mechanistic behaviour, and music structure. 
%
%
Therefore we seek covariance functions that can describe or reflect these properties. 

%
One powerful technique for constructing new kernels  is to build them out of simpler kernels as building blocks \cite{shawe04,bishop06}. Two useful properties we can use to build valid kernels are as follows:
%
$
k(t,t') = \phi(t) k_{1}(t,t') \phi(t'),
$
and
$
k(t,t') = k_{1}(t,t') + k_{2}(t,t'),
$
%
where $\phi(\cdot)$ is any function. Other properties can be found in \cite{bishop06}. We use these properties for building non-stationary covariance functions.  

\subsubsection{Kernels for describing non-stationarity}

To construct non-stationary kernels we combine basic stationary covariance functions.
 We use \textit{change-windows} in order to be able to model notes or sound events which are not continuously active but have a beginning and an ending in the music signal. As in \cite{lloyd2014} we define a change-window by multiplying two sigmoid functions,
that is
\begin{align}
\phi(t) 
=
\frac{1}{1 + e^{-\varsigma(t - \alpha)}   } \times \frac{1}{1 + e^{-\varsigma(\beta - t)} },
\end{align}
where $\varsigma$ determine how fast the change-window rises to its maximum value  or falls to zero, whereas  $\alpha,\beta$ defines the onset and the offset of the change-window respectively. 
%
%
The parameters of the change-windows are directly related with the location, onset and offset of the sound events we want to model.
In the present work we will use manually-specified onset/offset locations.
%

\begin{figure}[]
\centering
\subfigure[ Two change-windows.]{\includegraphics[width=0.5\columnwidth]{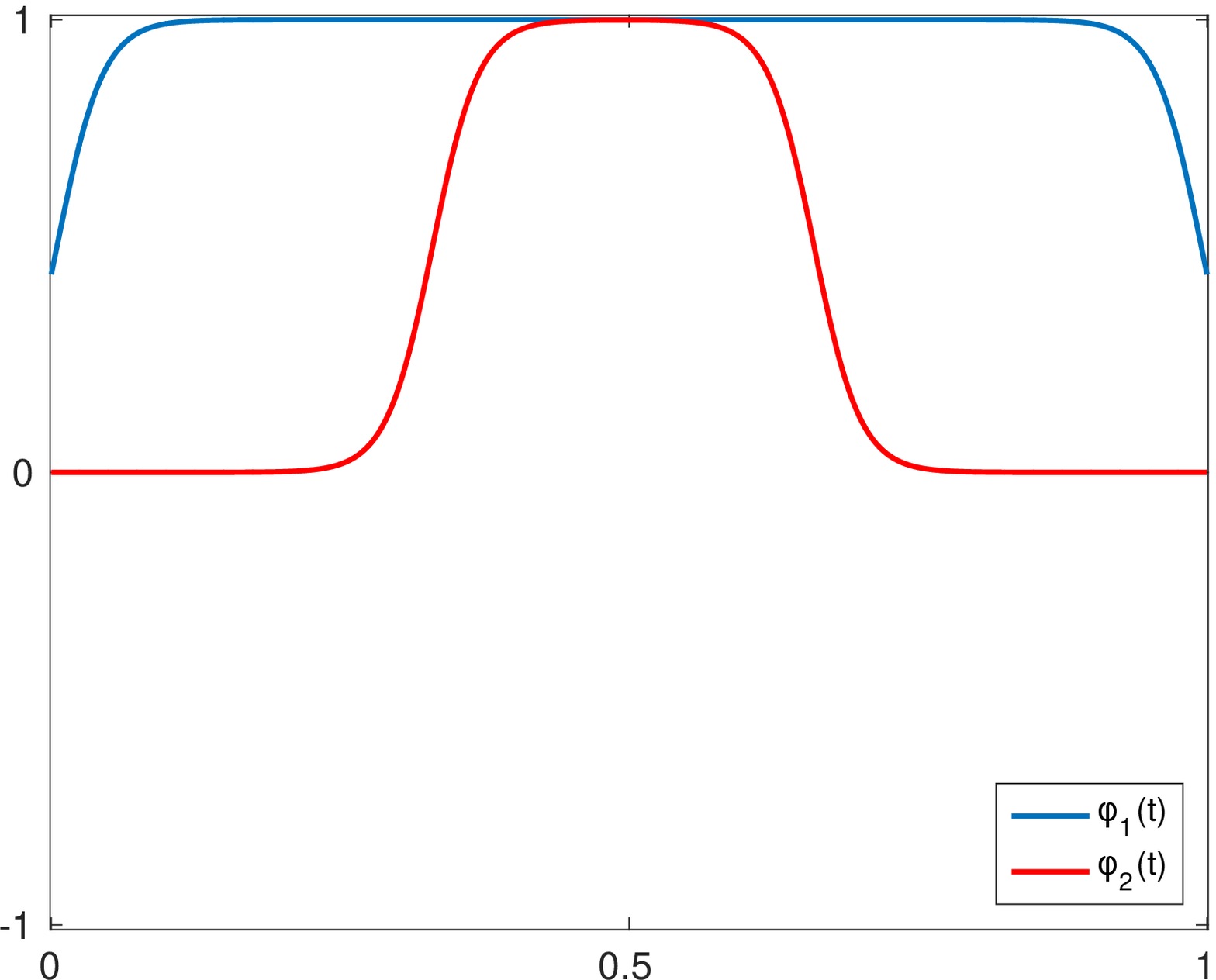}\label{f.change_win}}%
\subfigure[ Two function realizations.]{\includegraphics[width=0.5\columnwidth]{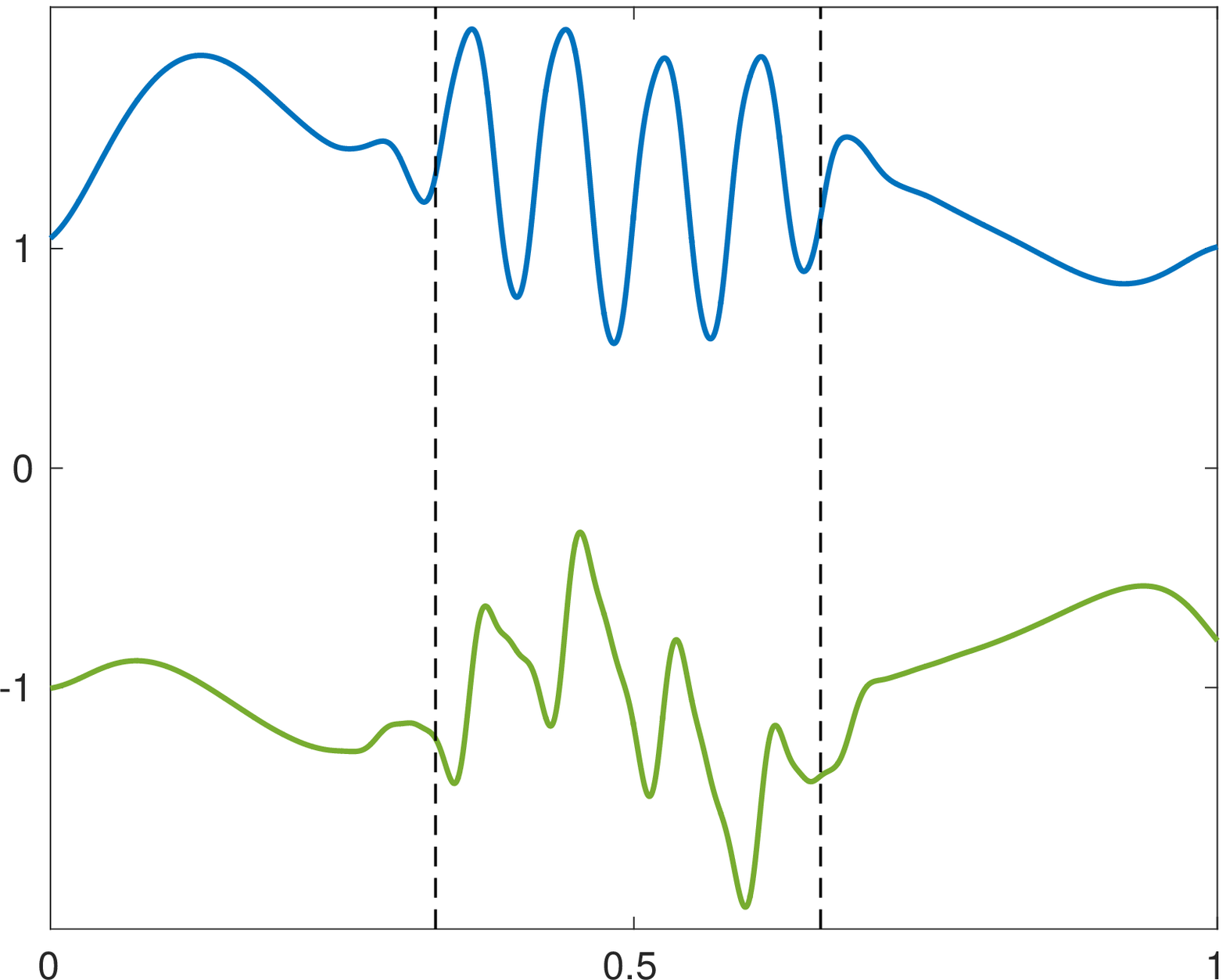}\label{f.two_samples}}%
\caption{Example of GP model with nonstationary kernel.}
\label{f.example}
\end{figure}

%
To illustrate the construction of a non-stationary kernel, an example model with two change-windows and its corresponding covariance functions is developed here. We assume  a GP $f(t) \sim \mathcal{GP}(0,k_{f}(t,t'))$ described by a linear combination of other two GPs $f_{i}(t) \sim \mathcal{GP}(0,k_{i}(t,t'))$ for $i = 1,2$, each one weighted by its corresponding change-window $\phi_i(t)$, i.e.
%
$
f(t) = \phi_{1}(t) f_{1}(t) + \phi_{2}(t)f_{2}(t).
$
%
%
%
%
%
%
%
Using the definition of covariance function \eqref{e.covgen} we can calculate the kernel $k_{f}(t,t')$ for the complete process $f(t)$ as follows:
%
%
%
$
k_{f}(t,t') =
  \phi_{1}(t)  k_{1}(t,t') \phi_{1}(t') +
  \phi_{2}(t)  k_{2}(t,t') \phi_{2}(t'),
$
%
where $k_{1}(t,t')$ and $k_{2}(t,t')$ are the covariance functions for the GPs $f_{1}(\cdot)$ and $f_{2}(\cdot)$ respectively.
%
%
%
%
This kernel configuration was used for generating the two samples shown in Figure \ref{f.two_samples}. 
In both cases $k_{1}(t,t')$ and $k_{2}(t,t')$ have the harmonic kernel form \eqref{e.kernel_ecq}  that we describe shortly (section \ref{timedynamics_ker_prop}).
%
%
%
Figure \ref{f.change_win} contains the shape of the two change-windows used ($\phi_1(t)$ blue, $\phi_2(t)$ red).
From Figure \ref{f.example} we see that with the proposed methodology we are able to describe non-stationary functions.
%

\subsubsection{Kernel general form for $M$ change-windows}
%
%
The previous example could be generalized for modelling the complete process $f(t)$ as a linear combination of  $M$ random process, representing each one a note or sound event. 
In this way
\begin{align}\label{e.general_f}
f(t) = \sum_{m=1}^{M} \phi_{m}(t) f_{m}(t),
\end{align}
where each Gaussian process $[f_1,f_2,\cdots,f_M]$ is independent with respect to each other. 
%
%
%
It is important to highlight that $M$ is directly related with the number of notes or sound events in the signal. On the other hand, $\phi_m(t)$ are respectively the change-windows or weight functions that allow a specific GP $f_m(t)$ to appear or vanish in certain parts of the input space (time). 
In this way, the general expression for the covariance function $k_f(t,t')$ is given by
%
%
\begin{align}\label{e.kernelgeneral}
k_f(t,t') 
=
 \sum_{m=1}^{M}  \phi_{m}(t) k_{m}(t,t') \phi_{m}(t').
\end{align}
We see that the overall process has a kernel consisting of a linear combination of the corresponding covariance functions of every subprocess.
%
%
%

\subsubsection{Rich spectral content kernel property}
\begin{figure}[]
\centering
\subfigure[$k_{\text{EQ}}(\tau)$]{\includegraphics[width=0.33\columnwidth]{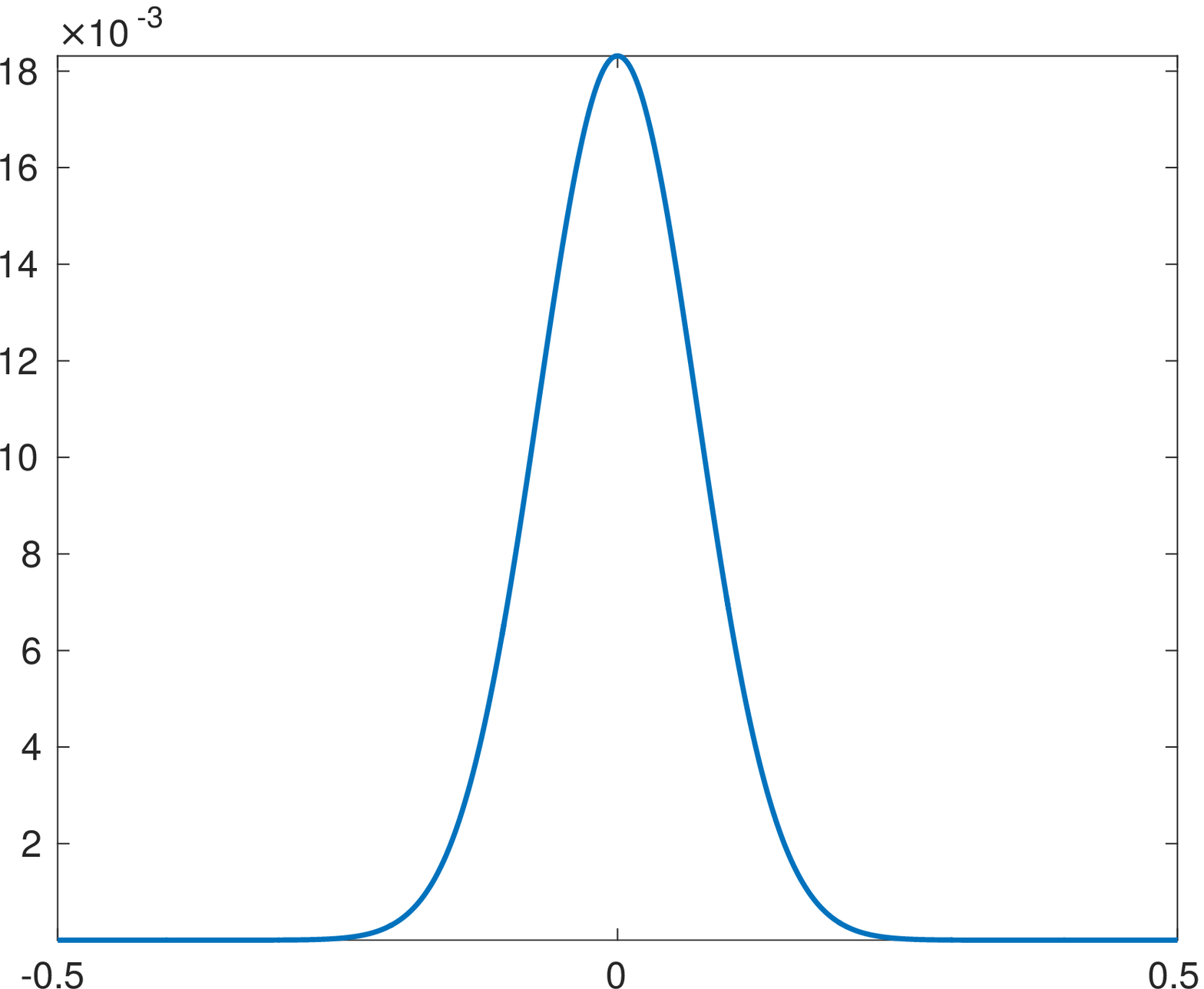}\label{f.kernel_expqua}}%
\subfigure[FT of $k_{\text{EQ}}(\tau)$]{\includegraphics[width=0.33\columnwidth]{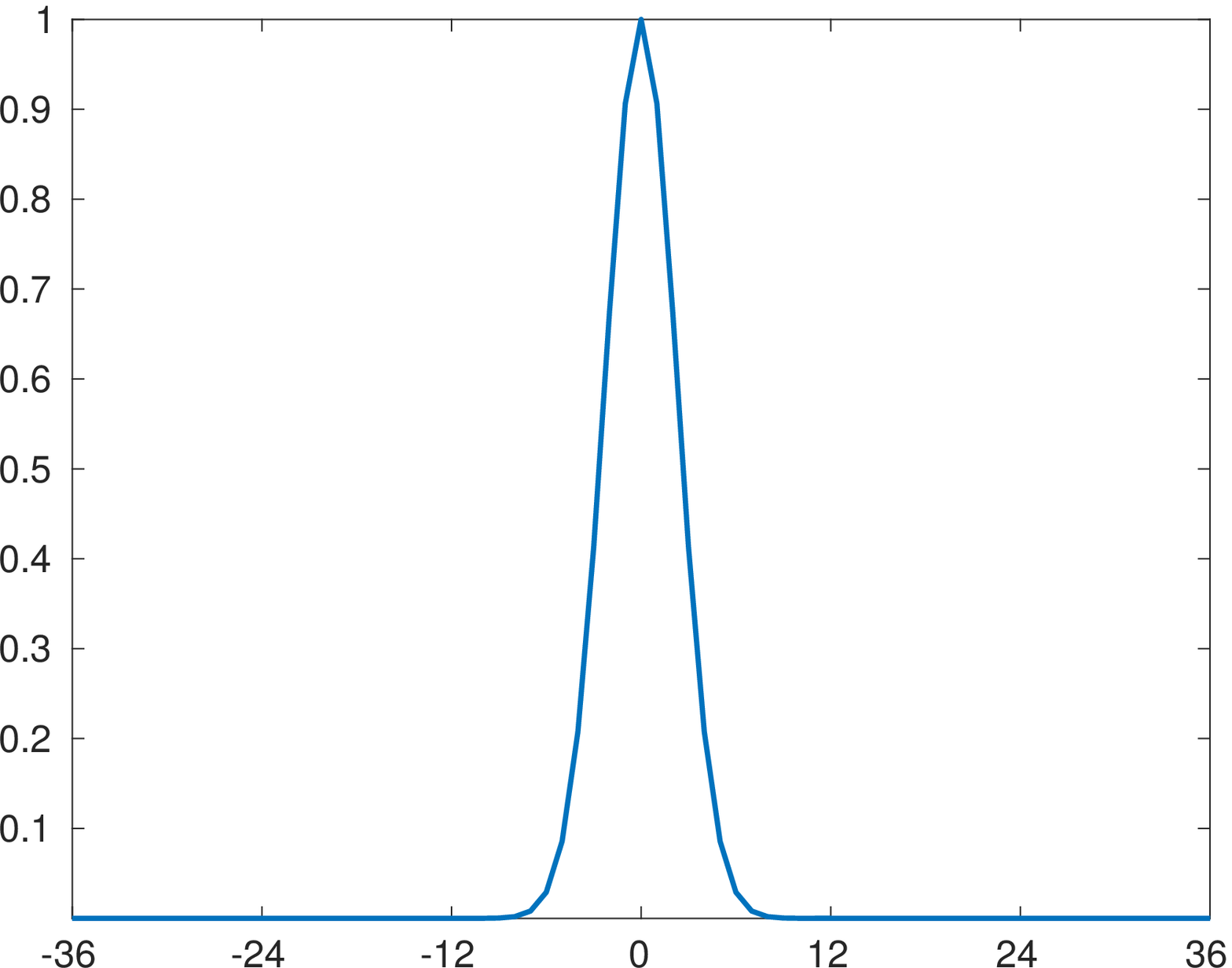}\label{f.FT_kernel_expcua}}%
\subfigure[Samples using $k_{\text{EQ}}$]{\includegraphics[width=0.34\columnwidth]{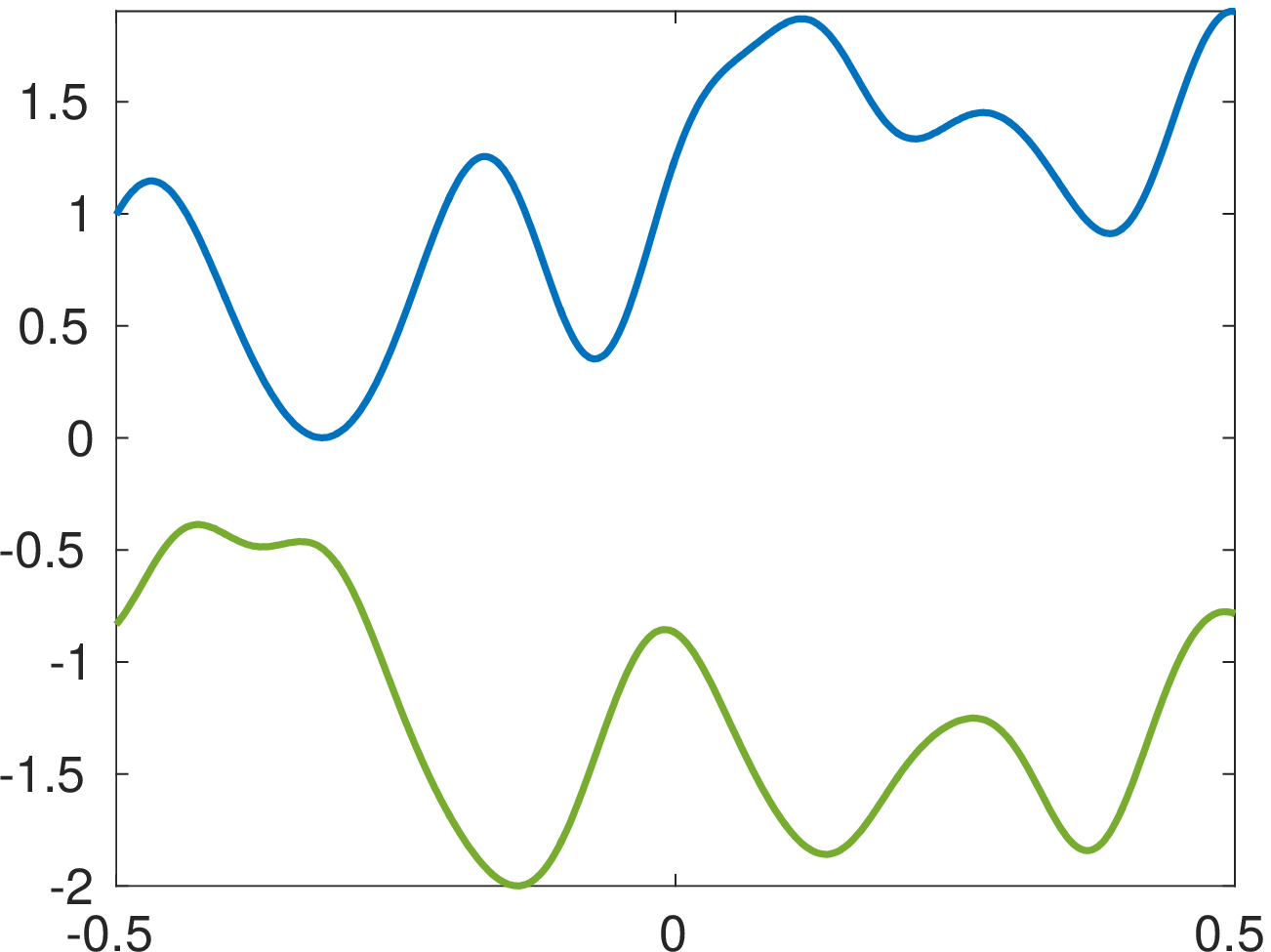}\label{f.samples_kernel_expqua}}%
\\
\subfigure[$k_{\text{EC}}(\tau)$]{\includegraphics[width=0.333\columnwidth]{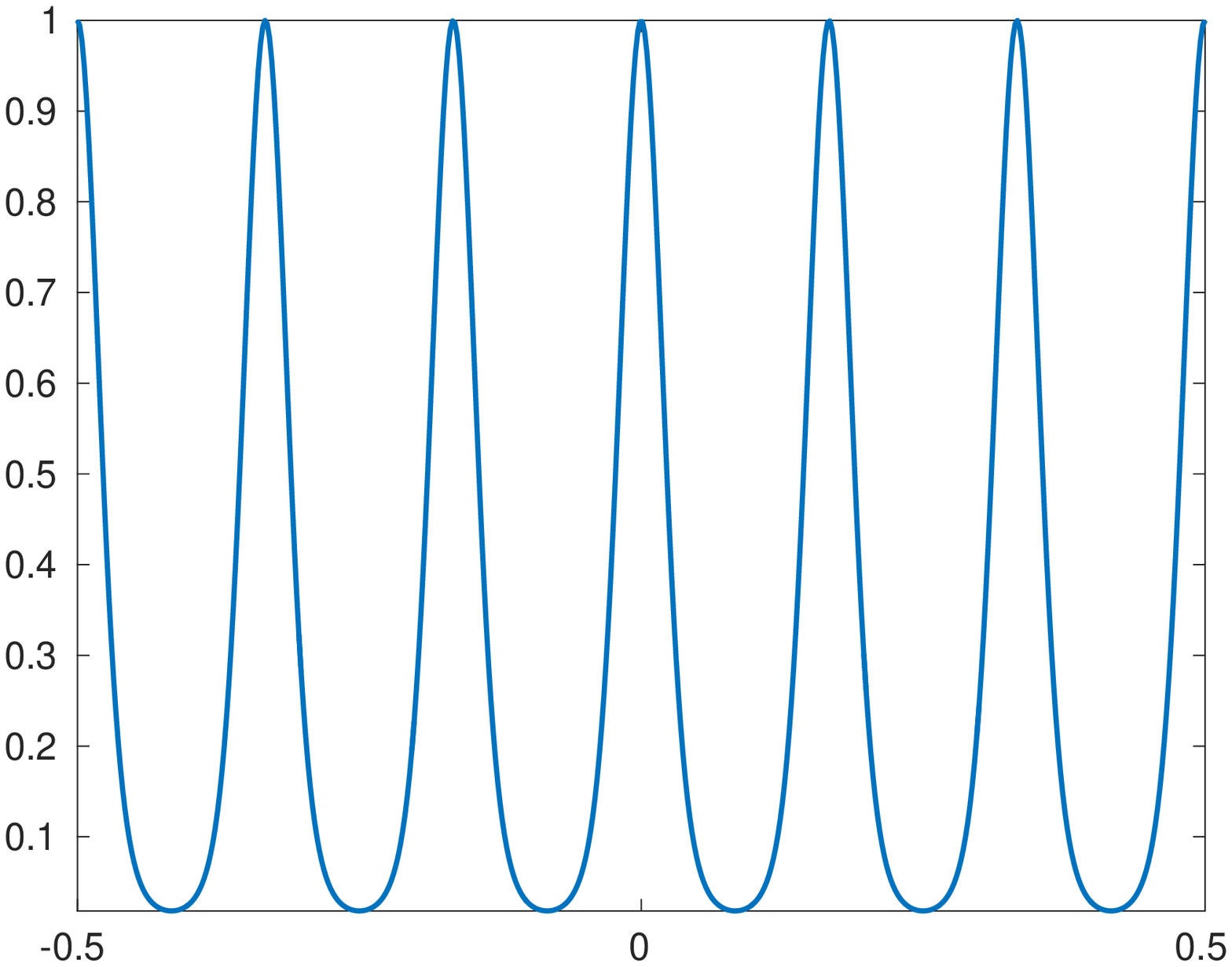}\label{f.kernel_expcos}}%
\subfigure[FT of $k_{\text{EC}}(\tau)$]{\includegraphics[width=0.333\columnwidth]{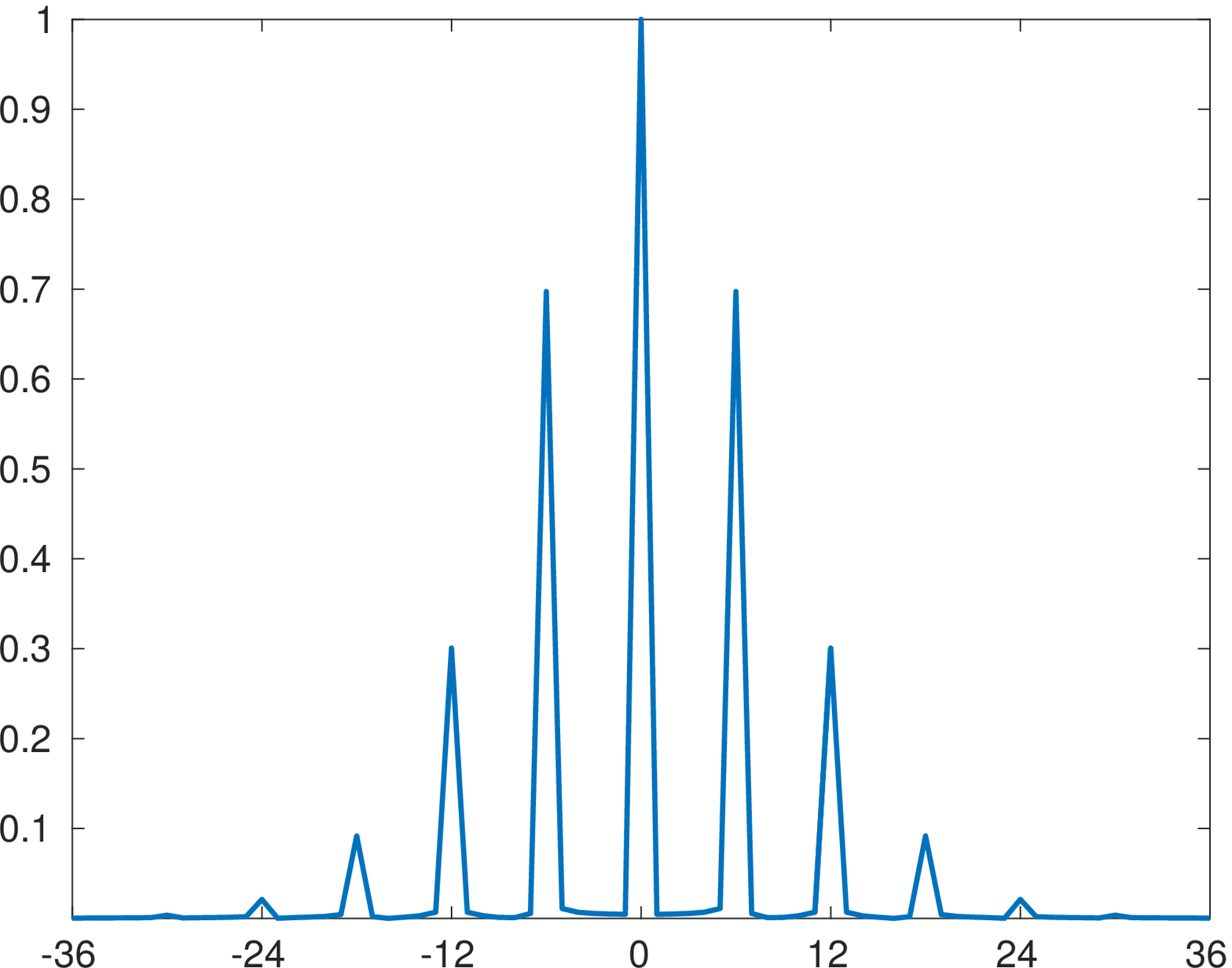}\label{f.fourier_kernel_expcos}}%
\subfigure[Samples using $k_{\text{EC}}$]{\includegraphics[width=0.333\columnwidth]{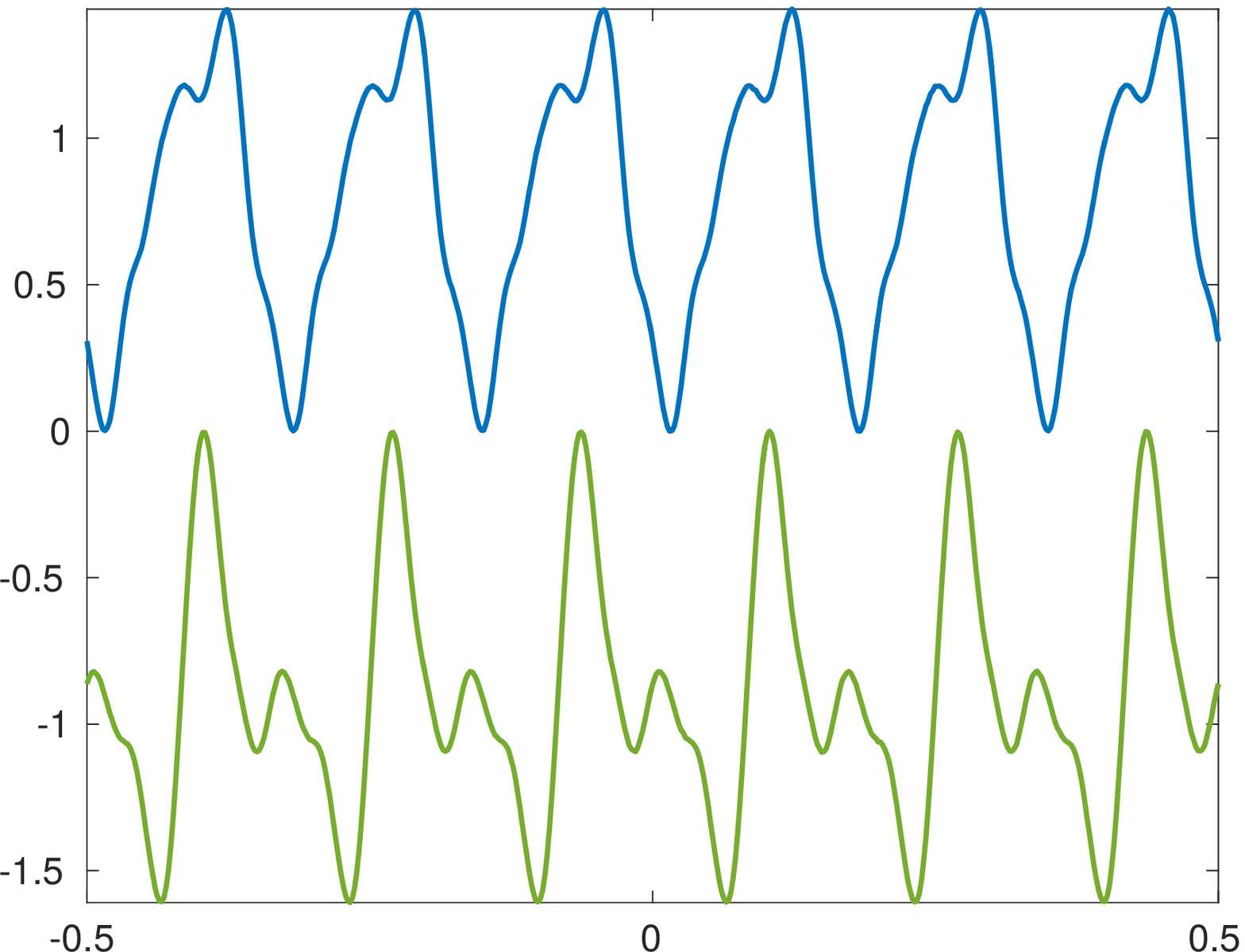}\label{f.samples_kernel_expcos}}%
\\
\subfigure[$k_{\text{ECQ}}(\tau)$]{\includegraphics[width=0.333\columnwidth]{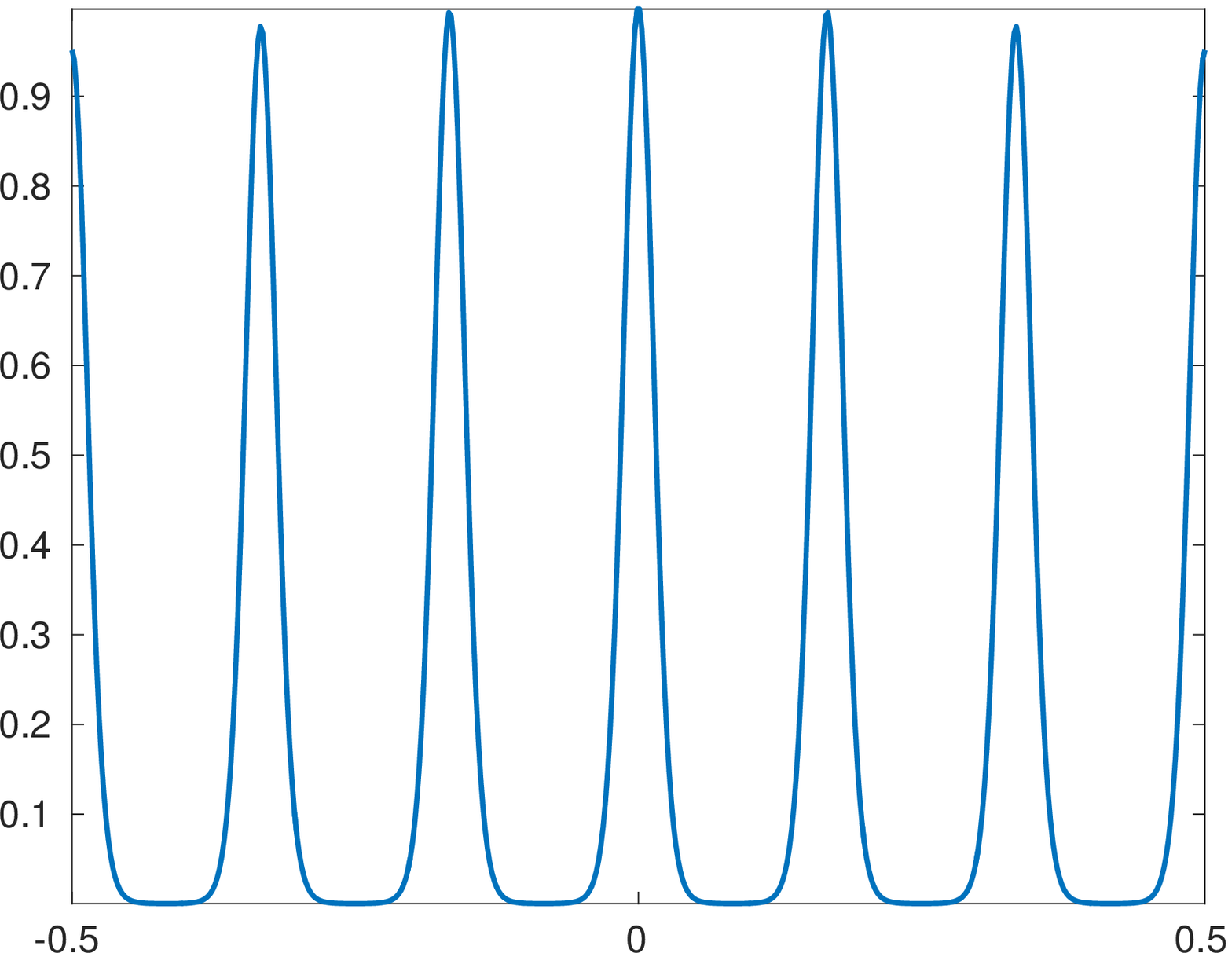}\label{f._kernel_expcosqua}}%
\subfigure[FT of $k_{\text{ECQ}}(\tau)$]{\includegraphics[width=0.333\columnwidth]{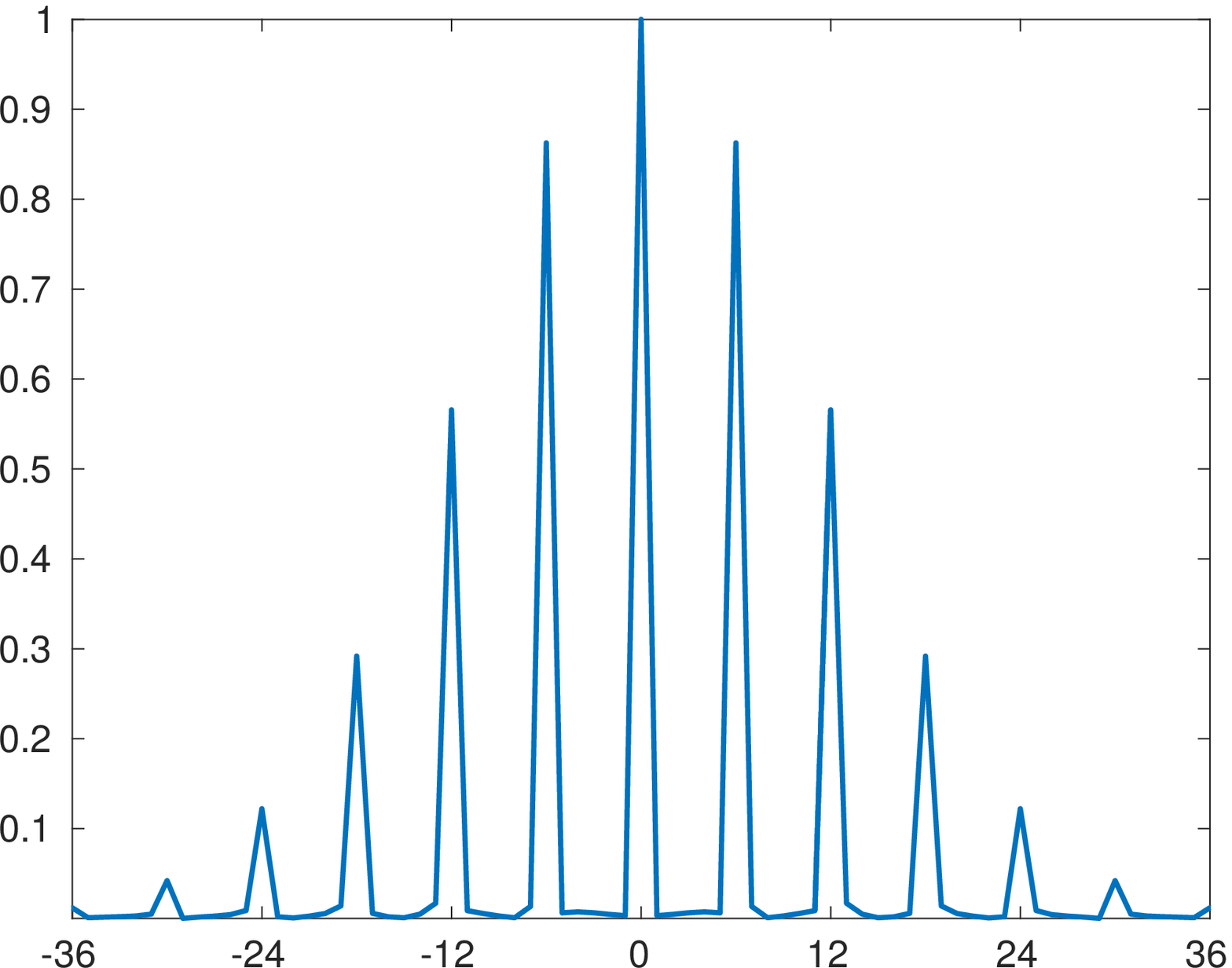}\label{f.fourier_kernel_expcosqua}}%
\subfigure[Samples using $k_{\text{ECQ}}$]{\includegraphics[width=0.333\columnwidth]{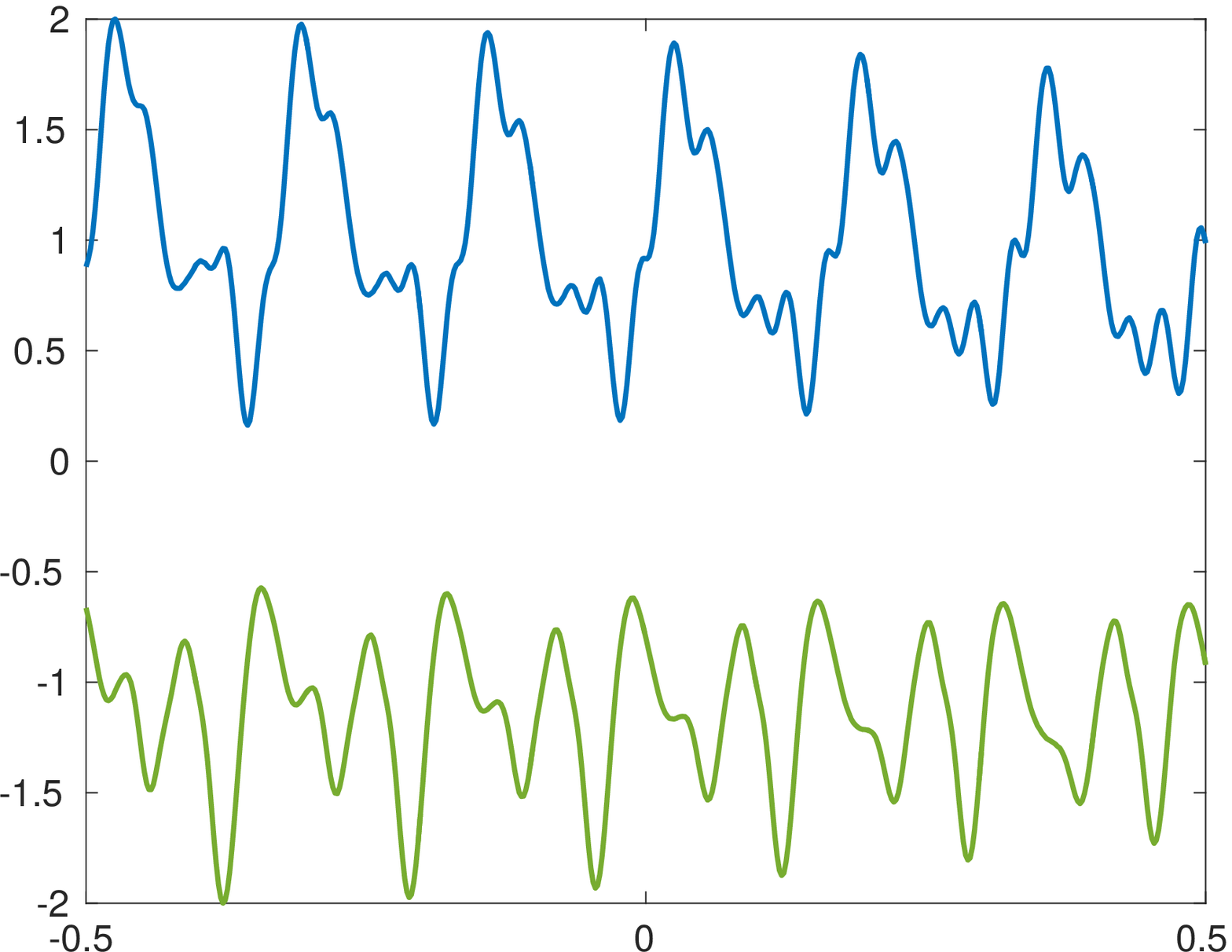}\label{f.samples_kernel_expcosqua}}%
\caption{
Frequency analysis of kernels \eqref{e.ker_expcua}, \eqref{e.kernel_expcos}, \eqref{e.kernel_ecq} depicted in (a), (d), (g) respectively. 
(b), (e), (h) are their corresponding FT. 
(c), (f), (i) show sampled functions.
}
\label{f.example_kernel_ecec}
\end{figure}

In the previous section we described the general form of the proposed kernel. Here we address the structure of the covariance functions $k_m(t,t')$ in \eqref{e.kernelgeneral}.
We assume that each GP  $f_m(t)$ in \eqref{e.general_f} is stationary. A random process is stationary (wide sense stationary WSS) if its mean is constant, and its kernel is a covariance function of $\tau = t-t'$, then we can write $k(t,t') = k(\tau)$  \cite{Shanmugan88,rasmussen05}.
%
%
%
It can be shown that the \textit{spectral density} or \textit{power spectrum} $S(s)$ of a WSS process corresponds to the Fourier transform (FT) of the covariance function, that is
\begin{align}
S(s) = \int_{-\infty}^{\infty} k(\tau) e^{-js\tau} \dif \tau,
\end{align}
thus
\begin{align}
k(\tau) = \frac{1}{2\pi}\int_{-\infty}^{\infty} S(s) e^{js\tau} \dif s.
\end{align}
This is known as the Wiener-Khintchine theorem \cite{rasmussen05,Shanmugan88}. Taking that into account, we can do frequency-domain analysis for several covariance functions and decide which kernel is more appropriate for modelling the spectral content of music signals. 
Again we illustrate these concepts by an example, we compare the FT of  the \textit{exponentiated quadratic} covariance function $k_{\text{EQ}}(\tau)$ with the FT of the \textit{exponentiated cosine} kernel $k_{\text{EC}}(\tau)$, that is
\begin{align}\label{e.ker_expcua}
k_{\text{EQ}}(\tau) &= \sigma^2 \exp\left( -\frac{\tau^2}{2 l^2}\right) ,
%
%
\\
\label{e.kernel_expcos}
k_{\text{EC}}(\tau) &= \sigma^2 \exp \left[ z \cos(\omega \tau) \right],
\end{align}
%
where to keep values up to one we set  $\sigma^2 = \exp(-z)$ in \eqref{e.kernel_expcos}, as well as $z = 2$, and $\omega = 2 \pi 6$. For \eqref{e.ker_expcua} $l=0.01$.
covariance function \eqref{e.ker_expcua} is probably the most widely-used kernel within the kernel machines field, because the GP with a exponentiated-quadratic covariance function is very smooth  \cite{rasmussen05}. Figures \ref{f.kernel_expqua} and \ref{f.kernel_expcos} represent the form of these kernels.
The spectral density of  the GP with kernel \eqref{e.ker_expcua} contains only low frequency components and does not have any harmonic structure (Figure \ref{f.FT_kernel_expcua}). 
Figure \ref{f.samples_kernel_expqua} shows two different realizations sampled from a GP with $k_{\text{EQ}}(\tau)$ kernel, these functions evolve smoothly without any periodic or harmonic properties.
On the other hand,
the spectral density (Figure \ref{f.fourier_kernel_expcos}) of the \textit{exponentiated cosine} kernel \eqref{e.kernel_expcos} presents a DC part, as well as components at the natural frequency $6$Hz and harmonics (integer numbers of $6$Hz). 
Figure \ref{f.samples_kernel_expcos} shows two functions sampled from a GP with covariance function \eqref{e.kernel_expcos}. These realizations present constant amplitude-envelope and periodic properties with a fundamental frequency together with several harmonics. 
%
%
%
%


\subsubsection{Time dynamics kernel property}\label{timedynamics_ker_prop}
In order to allow periodic kernels to describe functions where the amplitude envelope changes in time, we introduce a modification of expression  \eqref{e.kernel_expcos}. To do so, we multiply the periodic kernel with \eqref{e.ker_expcua}. The resulting covariance function, called \textit{exponentiated-cosine-quadratic} corresponds to 
%
%
%
\begin{align}\label{e.kernel_ecq}
k_{\text{ECQ}}(\tau) = \sigma^2 \exp\left[ z \cos(\omega \tau) -\frac{\tau^2}{2 l^2}\right],
\end{align}
where we assume $\sigma^2 = \exp(-z)$, $\omega = 2 \pi 6$, $z=5$ and $l = 4$.
%
%
%
%
Figure \ref{f.fourier_kernel_expcosqua} depicts the FT of \eqref{e.kernel_ecq}. 
We see that its spectral density keeps similar to the one obtained for \eqref{e.kernel_expcos} (see Figure \ref{f.fourier_kernel_expcos}). But the realizations sampled from a GP with this covariance function (Figure \ref{f.samples_kernel_expcosqua}) show a smooth variation in the amplitude envelope, and also maintain the properties described by  the previous kernel \eqref{e.kernel_expcos},  i.e. a periodic structure with natural frequency and harmonics.  
This covariance function \eqref{e.kernel_ecq} seems to be more appropriate for modelling music signals in comparison with the two kernels presented previously (\eqref{e.ker_expcua}-\eqref{e.kernel_expcos}). 

\section{Empirical Evaluation}
Experiments were done over real audio. We evaluated different kernel configurations on a pitch estimation task, and on a missing data imputation task.  
All experiments assume we previously know the number of change-windows and its locations. In the pitch estimation task all the parameters of the covariance function are known, except those related with the fundamental frequency of each sound event, i.e. the value of $\omega_m$ in  \eqref{e.kernel_expcos} and \eqref{e.kernel_ecq} when using these kernels in the general model \eqref{e.general_f}. Thus, we focus on optimizing only these model hyperparameters from the data. In the  missing data imputation task the score of the modelled piece of music audio is used for tuning manually the model hyperparameters. 
%
%

\subsection{Data}

\begin{figure}[]
\centering
\subfigure[Signal used for pitch estimation.]{\includegraphics[width=1\columnwidth]{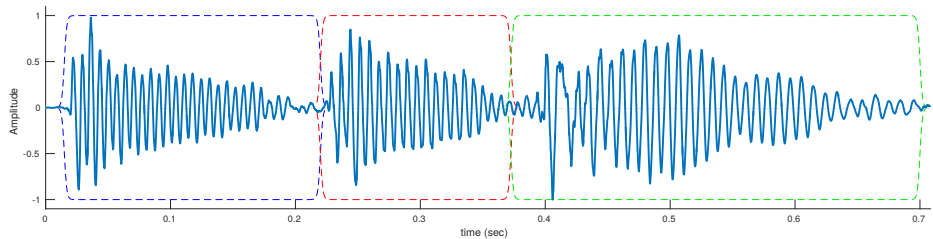}\label{f.data_pitch}}
\\
%
%
\subfigure[Signal used for filling missing-data gaps.]{\includegraphics[width=1\columnwidth]{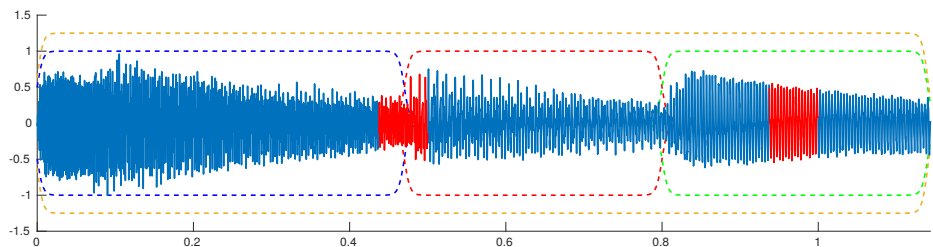}\label{f.data_gaps}}
\caption{(a) analysed audio (blue line), change-windows (dashed lines). (b) observed data (blue line), missing-data gaps (red line), change-windows (dashed lines).}
\end{figure}

In this study we used two short audio excerpts, in order to explore the method, so that we can efficiently fit models and search in the hyperparameter space.
The excerpt used for pitch estimation experiments corresponds to $0.7$ seconds of the song \textit{Black Chicken 37} by Buena Vista Social Club. This segment of audio contains three notes of a bass melody (Figure \ref{f.data_pitch}). 
In the missing data imputation task we used polyphonic audio corresponding to $1.14$ seconds of Chopin's \textit{Nocturne Op. 15 No. 1}, where more than one note occur at the same time. The segments of signal in red in Figure \ref{f.data_gaps} represent gaps of missing data.  
We reduced the sample frequency of both audio excerpts  from $44.1$KHz to $8$KHz.
%



\subsection{Hyper-parameter estimation}
To infer the hyper-parameters we consider an empirical Bayes approach, which allow us to use continuous optimization methods. We maximize the marginal likelihood. This moves us up one level of the Bayesian hierarchy, and reduces the chances of overfitting \cite{murphy12}.
%
%
%
%
%
%
Given an expression for the log marginal likelihood and its partial derivatives, we can estimate the kernel parameters using any standard gradient-based optimizer \cite{murphy12}.
A gradient descent method was used for optimization.

\section{Results and Discussion}
%

\subsection{Pitch estimation}
For the pitch estimation task we tested two different models with  kernels \eqref{e.kernel_expcos}, and \eqref{e.kernel_ecq} respectively. We performed hyperparameters learning using all the observed signal. This is because in this experiment rather than evaluating the prediction of the trained models, we were interested in the accuracy  of pitch estimation. Covariance function \eqref{e.ker_expcua} does not have any parameter we can link to the fundamental frequency of each sound event, that is why we omitted it here.
%
\subsubsection{Results using $k_{\text{EC}}(\tau)  $}

\begin{figure}[]
\centering
\subfigure[Observations (dots), and posterior mean (continuous line) using \eqref{e.kernel_expcos}.]{\includegraphics[width=1\columnwidth]{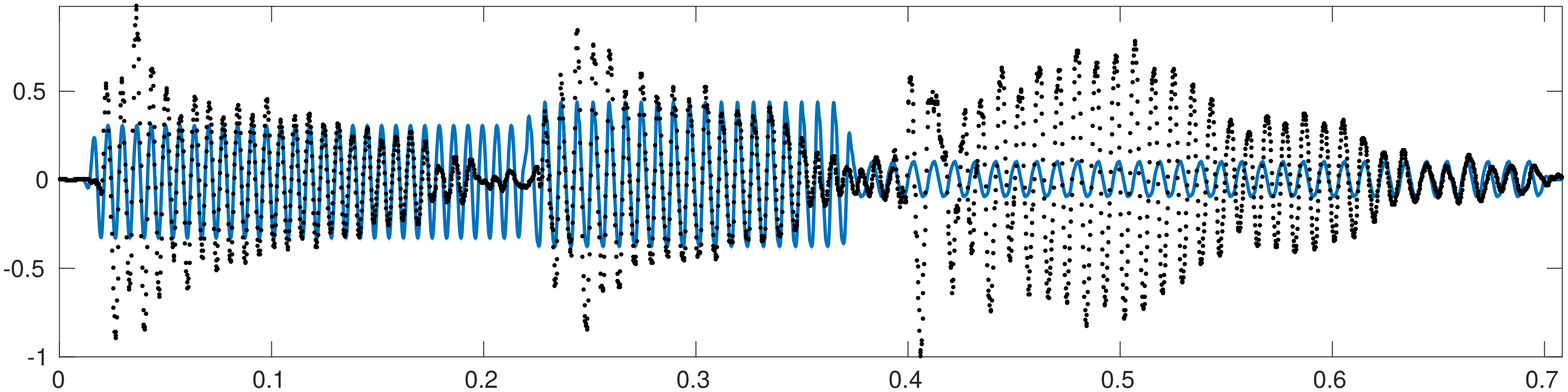}\label{f.experiment1_ec}}%
\\
\subfigure[Observations (dots), and posterior mean (continuous line) using \eqref{e.kernel_ecq}.]{\includegraphics[width=1\columnwidth]{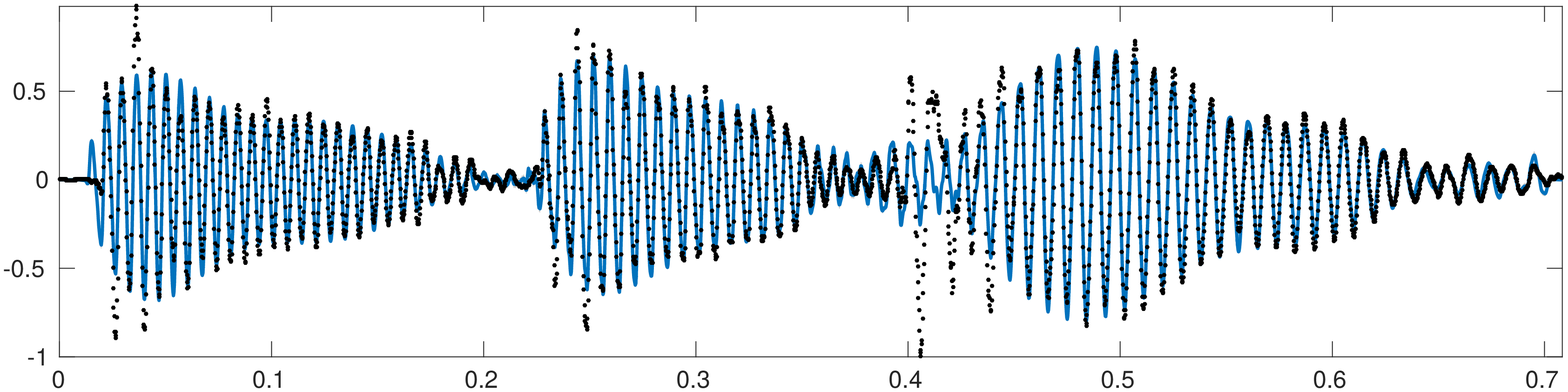}\label{f.experiment1_ecq}}%
\caption{Posterior mean for the pitch estimation experiments. (a) using $k_{\text{EQ}}(\tau)$, and (b) using $k_{\text{EQC}}(\tau)$.}
\end{figure}

We performed regression on the signal shown in Figure \ref{f.data_pitch} using the kernel \eqref{e.kernel_expcos}.
Figure \ref{f.experiment1_ec} shows the posterior mean of the predictive distribution after training (blue continuous line). The black circle points correspond to observed data. 
%
We see the trained model is able to estimate the pitch for each sound event with a RMS error of $0.6282$ semitones (Table \ref{t.experiment1}).
On the other hand, the amplitude-envelope evolution of the signal is beyond the scope of the structure that this kernel can model. This is because this covariance function can only describe constant amplitude-envelope, periodic signals, with a fundamental frequency and several harmonics (see Figure \ref{f.samples_kernel_expcos}).

%

\subsubsection{Results using $ k_{\text{ECQ}}(\tau) $}
To face the issue of modelling time dynamics we modified the previous covariance function \eqref{e.kernel_expcos}, by multiplying it with an exponentiated quadratic kernel \eqref{e.ker_expcua}. This allows to ``smooth" the strictly periodic behaviour of \eqref{e.kernel_expcos}. The resulting kernel corresponds to \eqref{e.kernel_ecq}.
From Figure \ref{f.experiment1_ecq} we see that although the posterior mean of the predictive distribution does not exactly fit the data, the model is able to learn the pitch of each of the three sound events with a smaller RMS error (Table \ref{t.experiment1}), as well as the time dynamics or variations in  the amplitude envelope of the signal.

\begin{table}[]
	\centering
	
	\caption{Pitch estimation RMS error (semitones).}
	\label{my-label}
	\begin{tabular}{|c|c|}
		\hline
		\textbf{kernel} & \textbf{RMS}  \\ \hline
		$k_{\text{EQ}}(\tau)$ & $--$   \\ \hline
		$k_{\text{EC}}(\tau)$ & $0.6282$  \\ \hline
		$k_{\text{EQC}}(\tau)$ & $0.1075$ \\ \hline
	\end{tabular}
	\label{t.experiment1}
\end{table}

\subsection{Filling gaps of missing data in audio}\label{section.gaps}
\begin{figure}[]
\centering
\subfigure[Prediction on transient gap using $k_{\text{EQ}(\tau)}$. ]{\includegraphics[width=0.5\columnwidth]{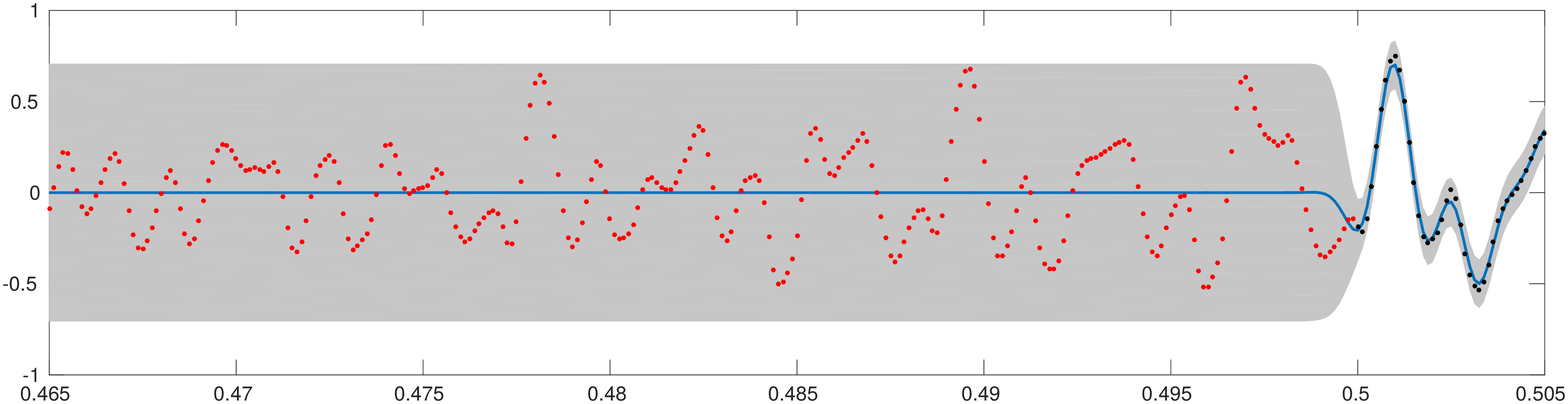}\label{f.experiment2_gap1_keq}}%
\subfigure[Prediction on smooth decay gap using $k_{\text{EQ}(\tau)}$.]{\includegraphics[width=0.5\columnwidth]{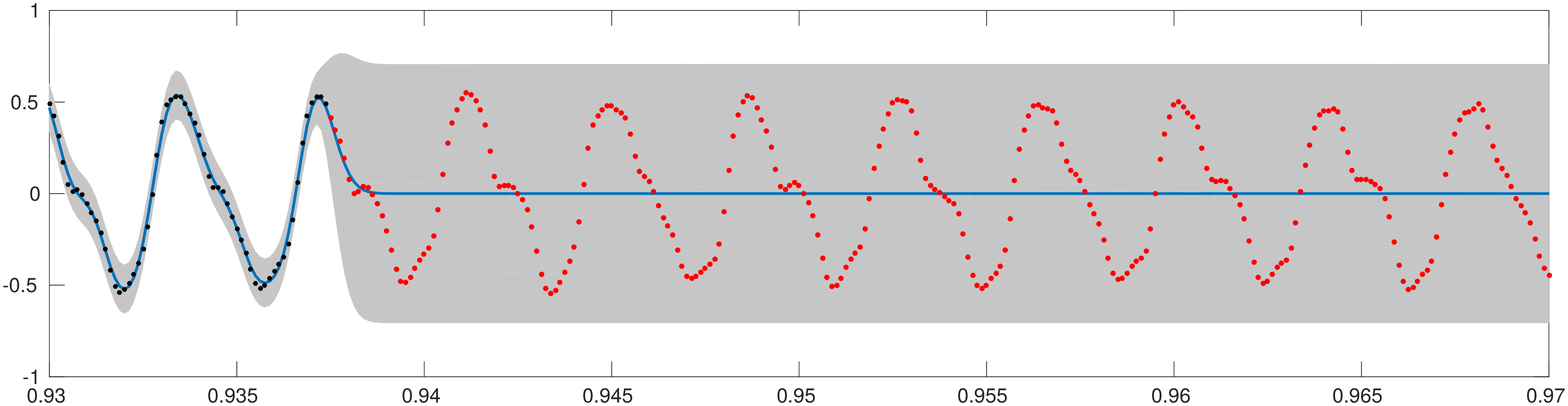}\label{f.experiment2_gap2_keq}}%
\\
\subfigure[Prediction on transient gap using $k_{\text{EC}(\tau)}$.]{\includegraphics[width=0.5\columnwidth]{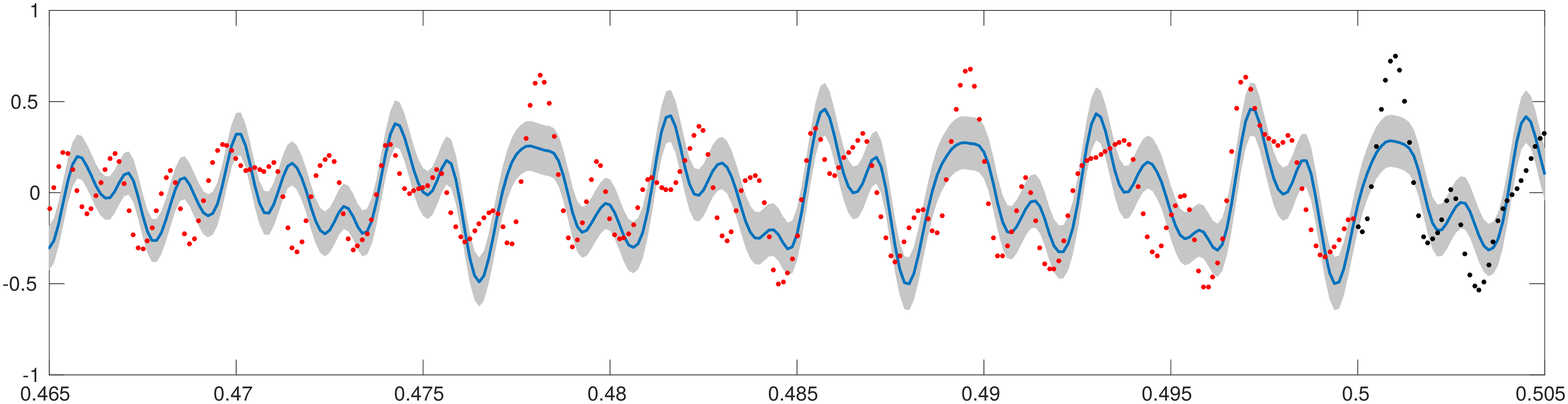}\label{f.experiment2_gap1_kec}}%
\subfigure[Prediction on smooth decay gap using $k_{\text{EC}(\tau)}$.]{\includegraphics[width=0.5\columnwidth]{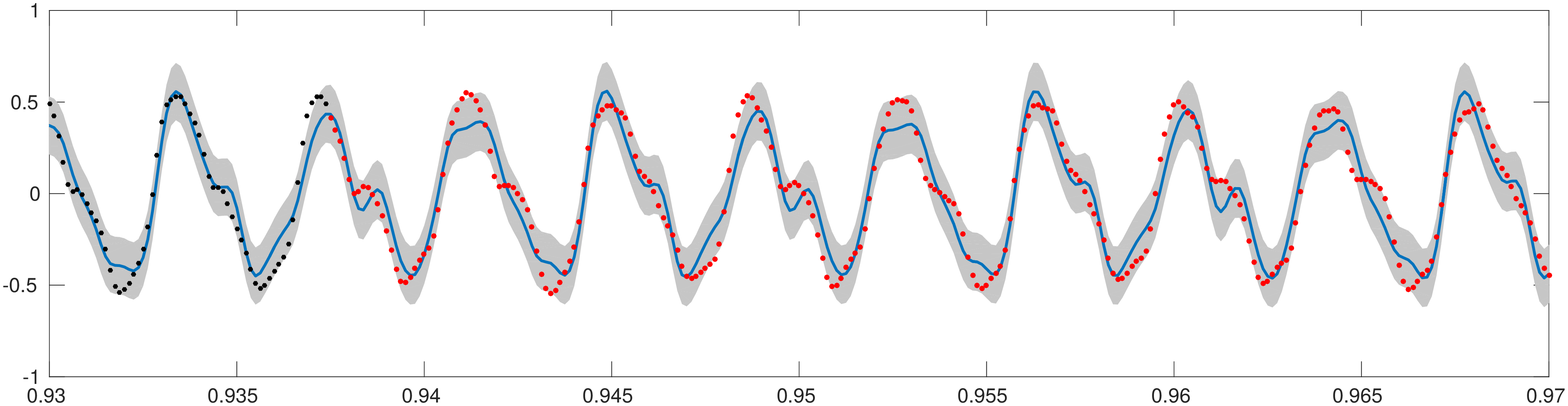}\label{f.experiment2_gap2_kec}}%
\\
\subfigure[Prediction on transient gap using $k_{\text{EQC}(\tau)}$.]{\includegraphics[width=0.5\columnwidth]{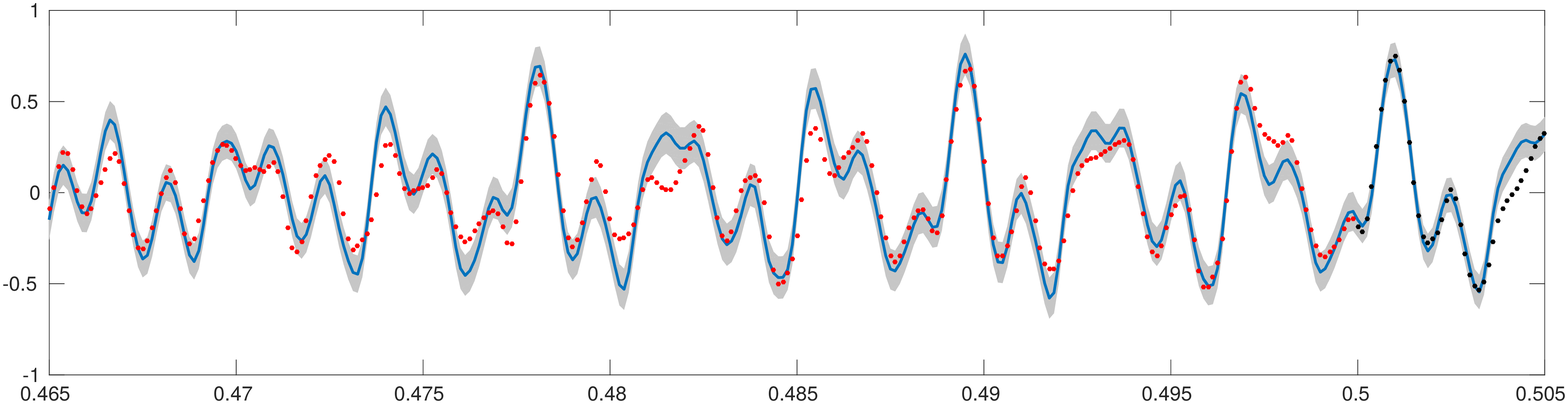}\label{f.experiment2_gap1_kecq}}%
\subfigure[Prediction on smooth decay gap using $k_{\text{EQC}(\tau)}$.]{\includegraphics[width=0.5\columnwidth]{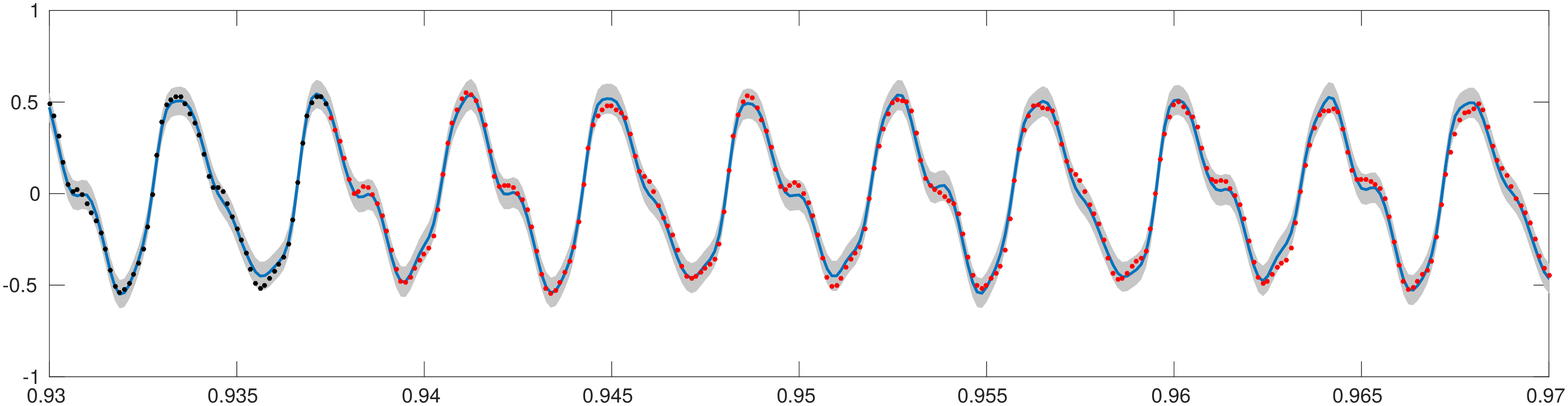}\label{f.experiment2_gap2_kecq}}%
\caption{Zoom in a portion of missing-data gaps. In each figure the continuous blue line represent the posterior mean, grey shaded areas correspond to the posterior variance, red dots are missing data, whereas black dots are observed data.}
\label{f.missingdata_expcua}
\end{figure}

%
%
%
%
We compared three different models predicting missing-data gaps. We studied kernels \eqref{e.ker_expcua}, \eqref{e.kernel_expcos}, and \eqref{e.kernel_ecq}.
In Figure \ref{f.data_gaps} first gap (red segment) contains the transient (onset and attack \cite{Bello05}) of a sound event, whereas the second gap is located in a more stable segment of the data (smooth decay).
Figures \ref{f.experiment2_gap1_keq}-\ref{f.experiment2_gap2_keq} depict the prediction using \eqref{e.ker_expcua}. These figures correspond to zoom in small sections of the signal where the gaps occur (Figure \ref{f.data_gaps}).
We see that the model using this kernel overfits the data, i.e. the posterior mean (blue line) fits all the observed data (black dots) with high confidence (grey shaded area), but the confidence decreases and the prediction is quite poor in the input space zones where the data is not available (red dots). Also, we see that the model using \eqref{e.ker_expcua} does not expect any periodic behaviour in the gaps. The RMS error for both gaps is presented in Table \ref{t.results2}.

%
%

Figures  \ref{f.experiment2_gap1_kec}-\ref{f.experiment2_gap2_kec} show the prediction using covariance function \eqref{e.kernel_expcos}. 
In the transient gap (Figure \ref{f.experiment2_gap1_kec}) the posterior mean (blue line) does not follows the data, this is because transients are short intervals during which the signal evolves in a nonstationary, nontrivial and unpredictable way \cite{Bello05}. opposite to  this, the model using kernel \eqref{e.kernel_expcos} can only describe the behaviour of constant amplitude-envelope periodic stochastic functions.
In the second gap (Figure \ref{f.experiment2_gap2_kec}) the posterior mean describes properly the periodic behaviour of the data, but it does not follow the amplitude-envelope of the observations.  This is because this covariance function is able to describe periodic functions that have several harmonic components. The drawback of this kernel is that it assumes constant the amplitude of the periodic stochastic functions that describes. These different performance on the prediction is reflected on the RMS error obtained for each gap (Table \ref{t.results2}).

Results using \eqref{e.kernel_ecq} are presented in
Figures \ref{f.experiment2_gap1_kecq}-\ref{f.experiment2_gap2_kecq}. We see that in Figure \ref{f.experiment2_gap2_kecq} the posterior mean describes properly the periodic behaviour and amplitude envelope smooth evolution of the modelled signal.
We observe that prediction on the decay gap using \eqref{e.kernel_ecq} is closer to the actual data (red dots) than the results obtained with \eqref{e.kernel_expcos} as well as \eqref{e.ker_expcua}. This is reflected in the smallest RMS error in table \ref{t.results2}. This is because \eqref{e.kernel_ecq} allows to describe periodic functions that have several harmonic components and time-varying amplitude envelope. 
On the other hand, the prediction performance reduces for the transient gap (Figure \ref{f.experiment2_gap1_kecq}). In order to model the onset, attack and decay of a sound event, covariance function \eqref{e.kernel_ecq} could be modified for modelling nonstationary amplitude envelope evolution.

\begin{table}[hb!]
	\centering
	\caption{Filling gaps prediction RMS error.}
	\label{my-label}
	\begin{tabular}{|c|c|c|}
		\hline
		\textbf{kernel} & \textbf{RMS  transient  gap} & \textbf{RMS decay gap} \\ \hline
		$k_{\text{EQ}}(\tau)$  & $0.2265$ & $0.3172$  \\ \hline
		$k_{\text{EC}}(\tau)$  & $0.2143$ & $0.0964$  \\ \hline
		$k_{\text{EQC}}(\tau)$ & $0.0912$ & $0.0355$  \\ \hline
	\end{tabular}
	\label{t.results2}
\end{table}

\subsection{Related work}\label{related_work}

In \cite{Turner14} GPs are used for time-frequency analysis as probabilistic inference. Natural
signals are assumed to be formed by the superposition of distinct time-frequency components, with the analytic goal being
to infer these components by applying Bayes' rule \cite{Turner14}.
GPs have also been used for underdetermined audio source separation. 
In \cite{Liutkus11} the \textit{mixture} signal is modelled as a linear combination of independent convolved versions of latent GPs or \textit{sources}. 
The model splits the \textit{mixture} signal in frames also considered independent, by using weight-functions. 
Thus each source is modelled as a series of concatenated locally stationary frames, each one with its corresponding covariance function.
With this assumption the resulting signal is supposed to be non-stationary \cite{Liutkus11}.
On the other hand, despite the approach we present also assumes the latent GPs $f_m$ in \eqref{e.general_f} as non-correlated, the observed signal is not framed into independent segments. 
Instead of using weight-functions that act over the observed data, we introduce change-windows $\phi_m$ influencing each latent GP ending up with latent processes representing specific sound events that happen at certain segments of time. 
Therefore the proposed model keeps the correlation between the observations throughout all the signal. That is what allows to make prediction in gaps of missing data (section \ref{section.gaps}). GPs have been used also for estimating spectral envelope and fundamental frequency of singing voice \cite{Yoshii13}, and for time-domain audio source separation \cite{Yoshii_ismir13}.

\section{Conclusions}
In this article we discussed a Gaussian processes regression framework for modelling music audio. We compared different models in pitch estimation as well as in prediction of missing data. We showed which kernels were more appropriate for describing properties of music signals, specifically: nonstationarity, dynamics, and spectral harmonic content. The advantage of this approach is that 
by designing a proper kernel we can introduce prior knowledge and beliefs about the properties of music signals, and use all that prior information to improve prediction.
The presented work could be extended using efficient representations of GPs in order to model larger audio signals.  
Other kernels could be studied, as the \textit{spectral mixture} for modelling harmonic content \cite{Wilson13}, and Latent Force models \cite{alvarez13} for describing mechanistic characteristics of the signal.

\bibliography{ismir2016biblio}
\bibliographystyle{plain}

\end{document}